\documentclass[letterpaper]{article} 
\usepackage{aaai25}  
\usepackage{times}  
\usepackage{helvet}  
\usepackage{courier}  
\usepackage[hyphens]{url}  
\usepackage{graphicx} 
\usepackage{natbib}  
\usepackage{caption} 
\usepackage{algorithm}
\usepackage{algorithmic}
\usepackage{multirow}
\usepackage{booktabs}
\usepackage{xcolor}
\usepackage{subcaption}
\usepackage{dsfont}
\usepackage{amssymb}
\usepackage{amsmath}
\usepackage{listings}
\DeclareCaptionStyle{ruled}{labelfont=normalfont,labelsep=colon,strut=off} 
\floatstyle{ruled}
\newfloat{listing}{tb}{lst}{}
\floatname{listing}{Listing}
\title{Specifying What You Know or Not for Multi-Label Class-Incremental Learning}
\author{
    Aoting Zhang\textsuperscript{\rm 1,3}, Dongbao Yang\textsuperscript{\rm 1,3}\equalcontrib, Chang Liu\textsuperscript{\rm 5}, Xiaopeng Hong\textsuperscript{\rm 4}\equalcontrib, Yu Zhou\textsuperscript{\rm 2}\\
}
\affiliations{
    \textsuperscript{\rm 1}Institute of Information Engineering, Chinese Academy of Sciences\\
    \textsuperscript{\rm 2}VCIP \& TMCC \& DISSec, College of Computer Science, Nankai University\\
    \textsuperscript{\rm 3}School of Cyber Security, University of Chinese Academy of Sciences\\
    \textsuperscript{\rm 4}Harbin Institute of Technology \\
    \textsuperscript{\rm 5}Tsinghua University\\
    \{zhangaoting, yangdongbao\}@iie.ac.cn, liuchang2022@tsinghua.edu.cn \\ hongxiaopeng@ieee.org, yzhou@nankai.edu.cn
}

\usepackage{bibentry}

\begin{document}

\maketitle

\begin{abstract}
Existing class incremental learning is mainly designed for single-label classification task, which is ill-equipped for multi-label scenarios due to the inherent contradiction of learning objectives for samples with incomplete labels. We argue that the main challenge to overcome this contradiction in multi-label class-incremental learning (MLCIL) lies in the model's inability to clearly distinguish between known and unknown knowledge. This ambiguity hinders the model's ability to retain historical knowledge, master current classes, and prepare for future learning simultaneously. In this paper, we target at specifying what is known or not to accommodate {H}istorical, {C}urrent, and {P}rospective knowledge for MLCIL and propose a novel framework termed as {HCP}. Specifically, (i) we clarify the known classes by dynamic feature purification and recall enhancement with distribution prior, enhancing the precision and retention of known information. (ii) We design prospective knowledge mining to probe the unknown, preparing the model for future learning. Extensive experiments validate that our method effectively alleviates catastrophic forgetting in MLCIL, surpassing the previous state-of-the-art by 3.3\% on average accuracy for MS-COCO B0-C10 setting without replay buffers.
\end{abstract}

%

\section{Introduction}
\hspace*{1em}\textit{To know what it is that you know, and to know what it is that you do not know--that is understanding.} 
\begin{flushright}
  ---The Analects
\end{flushright}

\begin{figure}[t]
  \centering
  \includegraphics[width=0.86\linewidth]{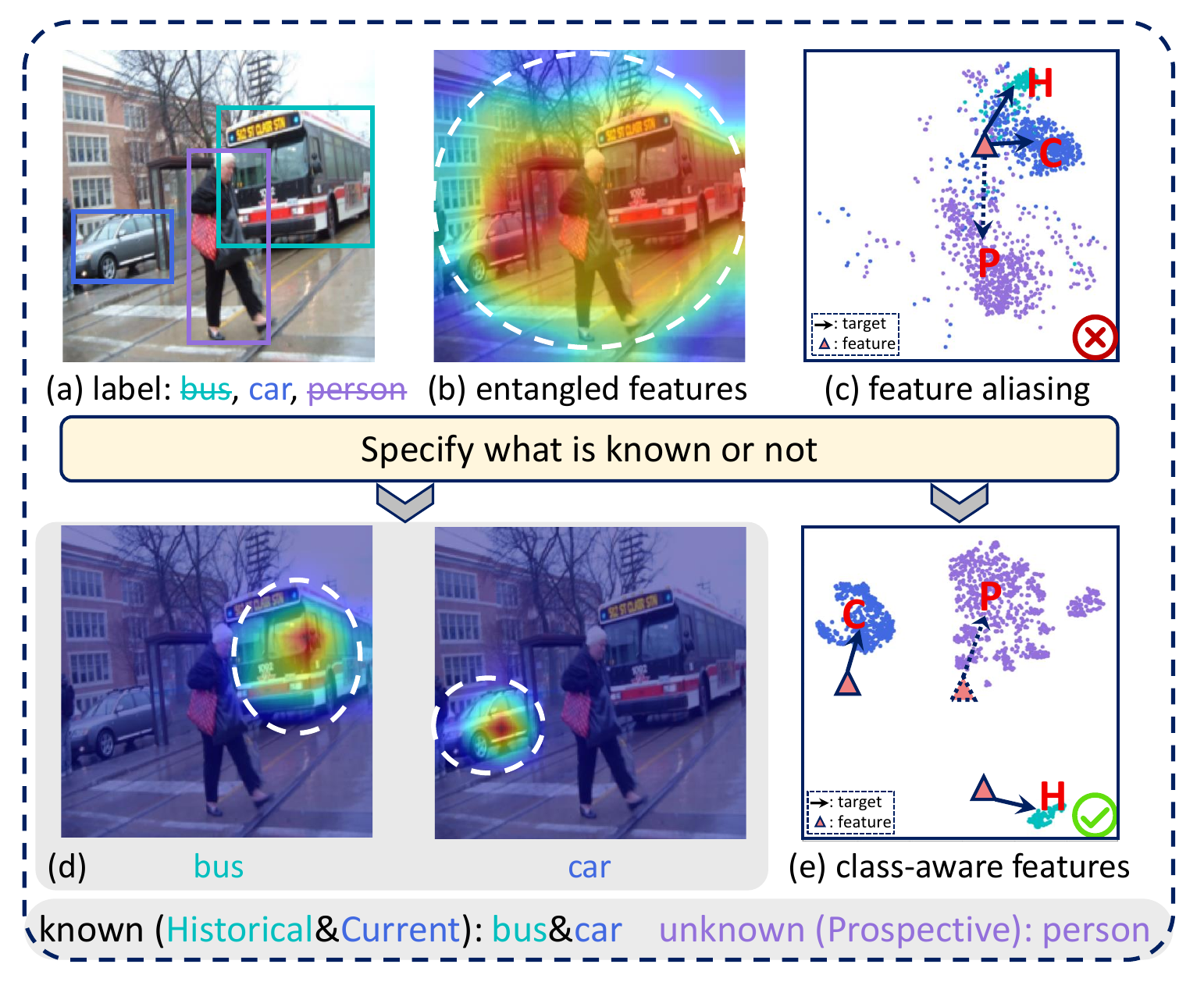}
  \caption{The contradiction of learning objectives in MLCIL arises from the model's inability to distinguish known and unknown knowledge. Current model fails to effectively recall prior known knowledge due to (a) the absence of historical labels, while (b) unknown classes' attention is inadvertently overlapped with known classes, and known classes are also entangled, resulting in (c) feature aliasing and contradictory learning objectives. By specifying what is known or not, (d) fine-grained class-aware features are focused, leading to (e) enhanced inter-class discriminability, alleviating the contradiction.}
  \label{fig:krt_ckat_attn}
\end{figure}
Class incremental learning (CIL)~\cite{aljundi2018memory, douillard2022dytox} is developed to continuously identify new classes while preserving old knowledge. Numerous studies endeavor to address the problem of catastrophic forgetting in CIL caused by the absence of old data. These studies are generally tailored to single-label class-incremental learning (SLCIL), assuming that each image only contains one single class. However, real-world images often feature multiple labels (\textit{e.g.}, a street scene depicts cars, buses, persons, etc.). To this end, multi-label class incremental learning (MLCIL) has caught progressive attention~\cite{dong2023knowledge}, aiming to correctly classify an image into multiple classes that may be introduced across different sessions.

Different from SLCIL, MLCIL typically involves images that simultaneously contain objects from historical, current and prospective classes since the foreground class definition evolves between incremental sessions. Annotations are available only for the classes learned at the current session (car in Figure~\ref{fig:krt_ckat_attn}), leaving past (bus) and future (person) classes unlabelled. 
Directly adopting anti-forgetting techniques in SLCIL, such as knowledge distillation and exemplar replay, yields poor results, as they fail to tackle the contradiction of learning objectives caused by incomplete labels inherent to MLCIL. Specifically, the learning objective contradiction arises from the inconsistency among three critical learning targets--preserving previously acquired knowledge, excelling in current classes, and preparing for future learning. Recent studies emphasize pseudo-labeling past classes and maintaining multiple knowledge to preserve previously acquired knowledge. For example, KRT~\cite{dong2023knowledge} proposes a knowledge restoration and transfer framework to address the issues of known-class label absence. Despite the substantial progress, these methods ignore another aspect, namely the interference of future classes that the model does not know at the current session. Without ``knowing'' future classes, the model inadvertently activates these features into known class representations. As shown in Figure~\ref{fig:krt_ckat_attn} (b), although the current model does not ``know'' person, it still pays high attention to the corresponding region and entangles the representations of person with known classes (car and bus). Meanwhile, due to the insufficient ``knowing'' ability, that is, fine-grained class-aware representation ability, past and current classes are also entangled in co-occurrence scenarios. Entangled features of historical, current and prospective classes blur the knowledge boundaries between tasks, resulting in the contradiction of learning objectives for aliased features. It not only impacts the learning for future classes but also degrades the recall of known knowledge due to the lack of target supervision, aggravating catastrophic forgetting.

In this work, we aim to specify what is known or not to accommodate Historical, Current, and Prospective knowledge for MLCIL and propose a novel framework termed as HCP. For clarifying known knowledge, HCP first proposes a dynamic feature purification module to capture fine-grained class-aware features, preventing feature aliasing across sessions. Specifically, each class is assigned a class embedding, activating relevant features within the image based on attention mechanism, and classes can be flexibly expanded by adding new embeddings. As shown in Figure~\ref{fig:krt_ckat_attn} (d), our method focuses the attention of the bus embedding entirely on the bus itself and eliminates other noises. HCP then effectively recalls old known knowledge through pseudo-labeling with distribution prior, which alleviates the problem of large forgetting differences between classes. For probing unknown knowledge, we mine knowledge from images that encompass historical, current, and prospective classes to develop features pertinent to future classes. This prospective strategy helps the model learn a richer feature set and clearly defines the boundaries of current classes, thereby enhancing the discriminability of current features and preparing the model for future learning.

{To summarize, our major contributions are as follows:}
\begin{itemize}
    \item We reveal the challenge of learning objectives contradiction in MLCIL, and propose a new framework named HCP, aiming to specify what is known or not to accommodate historical, current, and prospective knowledge.
    \item For clarifying the known, we develop dynamic feature purification that focuses on fine-grained class-aware features of known classes. In addition, we design recall enhancement with distribution prior to effectively preserve old known knowledge.
    \item For probing the unknown, we mine knowledge to develop features pertinent to future classes, boosting the model’s discriminative capacity and preparing for future learning.
    \item Experiments on various settings demonstrate that our method achieves state-of-the-art performance and effectively mitigates catastrophic forgetting in MLCIL.
\end{itemize}

\section{Related Work}
\noindent \textbf{Single-Label Incremental Learning}
aims to integrate new concepts without forgetting previously learned~\cite{zhu2025reshaping}. Current mainstream methods are typically divided into three categories. \textit{Regularization-based methods} design a loss function to penalize changes in the weights or activations during learning new tasks~\cite{schwarz2018progress, huang2024etag, yang2022multi}. 
\textit{Rehearsal-based methods} involve retaining a subset of previously encountered samples and merging them with new data for training~\cite{bang2021rainbow, chaudhry2018efficient, huang2024kfc}. Although they show impressive results, relying on a memory buffer raises concerns about the privacy of stored images and increases storage space.
\begin{figure*}[t]
  \centering
  \includegraphics[width=0.86\linewidth]{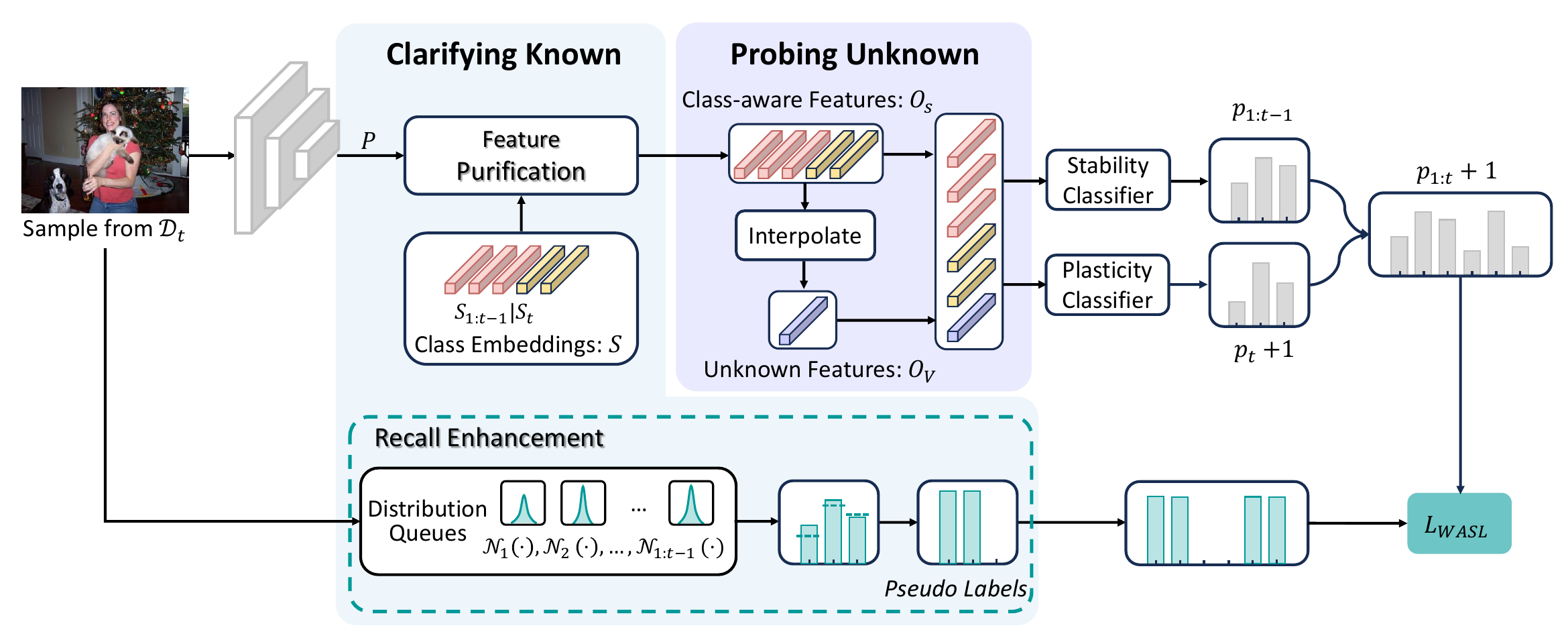}
  \caption{Framework of HCP, which leverages Clarifying Known and Probing Unknown to accommodate historical, current, and prospective knowledge. For clarifying known knowledge, we design dynamic Feature Purification to capture fine-grained class-aware features $O_{s}$ to avoid feature aliasing across sessions, and Recall Enhancement with distribution prior to effectively retain historical known knowledge. For probing unknown knowledge, we interpolate known features as prospective class to help enrich the feature set, enhancing the discriminability of known features and facilitate future learning.}
  \label{framework}
\end{figure*}
\textit{Architectural-based methods} modify the network architecture by adding sub-networks or experts when new tasks arrive while keeping the previous network frozen~\cite{douillard2022dytox, wang2022learning}.  

\noindent \textbf{Multi-Label Classification}
has been a challenging problem compared with single-label classification. The straightforward approach is to treat each category independently and formulate it as multiple binary classification. However, it ignores the label correlation and spatial dependency between objects, which is important for multi-label. Therefore, several works~\cite{wang2016cnn, chen2018order} use the recurrent neural network (RNN) to capture label correlation, which face difficulty in parameter optimization. Others apply Graph Convolutional Network (GCN)~\cite{zhou2020graph} to model label relationships~\cite{you2020cross, chen2019multi}, which capture spurious correlations when the label statistics are insufficient. Considering the binary cross entropy loss often suffers from the positive-negative imbalance issue, asymmetric loss (ASL)~\cite{guo2019visual} is designed to dynamically down-weights and hard-thresholds easy negative examples. Recently, some approaches~\cite{zhu2022two, li2023patchct, zhu2023scene} utilize transformer to model label correlation and improve multi-label prediction.

\noindent \textbf{Multi-Label Class-Incremental Learning}
has gained widespread attention with the rapid development of class incremental learning and multi-label classification. PRS~\cite{kim2020imbalanced} proposes a new sampling strategy for replay to alleviate the impact of imbalanced class distribution in buffers. OCDM~\cite{liang2022optimizing} leverages greedy algorithm to update memory quickly and efficiently. AGCN~\cite{du2023multi} utilizes a GCN network to build stable relationships between labels. KRT proposes a knowledge restore and transfer framework to solve the label absence of old classes. Although KRT achieves SOTA performance, the feature aliasing problem is still yet to be resolved, and the efficiency of KRT is impacted since it needs to go through the decoder multiple times to obtain task-level representation of each session. In this work, in addition to preserving previously acquired knowledge, we focus on accommodating historical, current and prospective knowledge simultaneously, overcoming the problem of learning target contradiction.

\section{Proposed Method}
\subsection{Problem Formulation}
The objective of MLCIL is to build a unified model that can recognize all encountered classes. We assume total $T$ sessions to simulate the continuous process. Given the dataset $D=\left \{ (x,y) \right \}$, where $x$ is the image with corresponding ground-truth label $y$, we split the dataset into $T$ subsets according to the lexicographical order of category names, with $\left \{ D_{1},\dots ,D_{T} \right \} $ and their label sets $\left \{ C_{1},\dots ,C_{T} \right \} $. All label sets are mutually exclusive, $i.e.$ $\forall m,n(m\ne n), C_{m}\cap C_{n}=\emptyset $. At session $t$, the model is trained on only $D_{t}$, with label space $C_{t}$. Different from SLCIL where each image has only one label, ground-truth $y$ in the multi-label scenario may contain classes from old sessions $C_{1:t-1}$ and future sessions $C_{t+1:T}$., so we preserve only the labels belonging to $C_{t}$ and discard others. During testing, the model is evaluated to recognize all seen classes $C_{1:t}=C_{1}\cup\dots \cup C_{t}$.

\subsection{Overview}
Our proposed framework HCP is illustrated in Figure~\ref{framework}. The key idea of HCP is to specify what is known or not at the current incremental session to accommodate historical, current, and prospective knowledge, thereby resolving learning target contradiction. Specifically, to clarify the known knowledge, the HCP framework initially introduces a dynamic feature purification module, where each class embedding focuses on fine-grained class-aware features without covering multiple classes, avoiding feature aliasing across sessions. It can flexibly introduce new classes in incremental learning by continuously adding new class embeddings. Moreover, we enhance the recall of historical knowledge by effectively utilizing priors of previous models, alleviating the problem of large forgetting differences between classes. To probe the unknown knowledge, we interpolate class features as a prospective class, which pushes all other non-target class features away from the generated component. As a result, the features of known classes are optimized to be more compact, facilitating future learning.

\subsection{Clarifying Known Knowledge}
In this section, we introduce feature purification module and how it adapts to multi-label incremental task. We then analyse the confidence forgetting between classes and further propose recall enhancement with distribution prior. 

\textbf{Feature Purification.}
To avoid feature aliasing between sessions caused by noise and impurity from non-target features, we propose feature purification module to extract fine-grained class-aware features from entangled multi-class global features. Compared with KRT, it ensures unique representations for each class and predicts historical and current classes in parallel. 

Given a sample from $D_{t}$, global features are first extracted and then reshaped to patch tokens $P\in \mathbb{R} ^{L \times d}$, where $L=h\cdot w$ and $h,w,d$ represent the height, width and dimension. To aggregate the object information and extract fine-grained class features, each class is assigned a learnable embedding and we get a sequence of class embeddings ${S}\in \mathbb{R} ^{M \times d}$, where $M=\left | C_{1:t} \right | $ is the number of known classes at session $t$. Feature purification module which consists of $L$ multi-head self-attention blocks, takes class embeddings $S$ and feature tokens $P$ as input to generate purified class features $O_{S}\in \mathbb{R}^{M \times d}$ and  enhanced patch features $O_{P}\in \mathbb{R}^{M \times d}$ (we omit the mini-batch):
\begin{align}
    (Q,K,V)&=(W_{q},W_{k},W_{v})[P,S], \\
    O &= W_{o}\mathrm{softmax}\left ( \frac{QK^{T}}{\sqrt{d/h} }  \right )V+ b_{o},
\end{align}
where $O=[O_{P},O_{S}]$ and $h$ is the number of attention heads. Based on the attention mechanism, each class embedding not only pays attention to the spatial information of global feature tokens, but also captures contextual relationships with other class embeddings. Then class features $O_{S}$ are fed to the classifier to obtain the output logits $p$.
\begin{figure}[t]
  \centering
\includegraphics[width=0.8\linewidth]{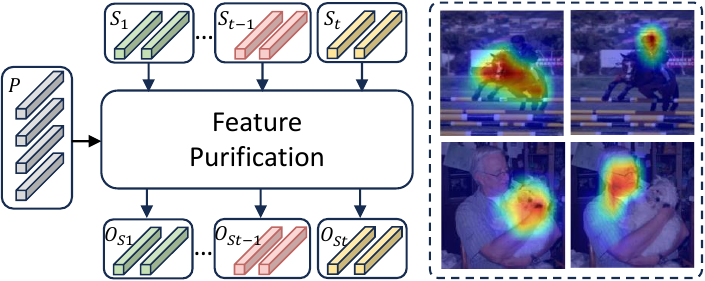}
  \caption{Illustration of feature purification. Each session appends new class embeddings $S_{t}$ for new class features $O_{St}$.}
  \label{fig: feature_purification}
\end{figure}

As shown in Figure~\ref{fig: feature_purification}, it can flexibly adapt to the learning of new sessions by appending new class embeddings, which is more efficient where old and new classes can be predicted in parallel. Following previous work~\cite{yan2021dynamically}, we utilize stability classifier to predict old logits $p_{1:t-1}$ and plasticity classifier to get new classification logits $p_{t}$, which are merged into a complete logits $p_{1:t}=[p_{1:t-1},p_{t}]$ for classification. During learning new classes, we freeze old embeddings $S_{1:t-1}\in \mathbb{R}^{|C_{1:t-1}|\times d}$ and stability classifier to maintain the old knowledge while new class embeddings $S_{t}\in \mathbb{R}^{|C_{t}|\times d}$ and plasticity classifier can adapt to new data. Before entering the next session $t+1$, the plasticity classifier and stability classifier are combined to form a new stability classifier, and then a new plasticity classifier is created for the new session.

\textbf{Recall Enhancement.}
Since the historical model has obtained effective information of old classes, we leverage the class probabilities $P=[p_{1},\cdots,p_{|1:C_{t-1}|}]$ predicted by the old model to recall historical knowledge~\cite{yang2022rd, yang2023one, yang2023pseudo}. The missing past labels $\hat{y}=[\hat{y}_{1},\hat{y}_{2},\cdots,\hat{y}_{|1:C_{t-1}|}]$ can be obtained: $\hat{y}_{k}=1$ if $p_{k}\ge  \varepsilon$ otherwise $0$, where $\varepsilon$ controls the quality of pseudo-labels.
However, category concepts present different learning difficulties for the model. Some classes have clear and easily distinguishable features therefore less-forgetting, while others may be subtle therefore more-forgetting. Treating all classes uniformly brings numerous noises, leading to the dilemma of either mislabeling false positive labels or neglecting true positive labels. 
\begin{figure}[t]
  \centering
  \includegraphics[width=\linewidth]{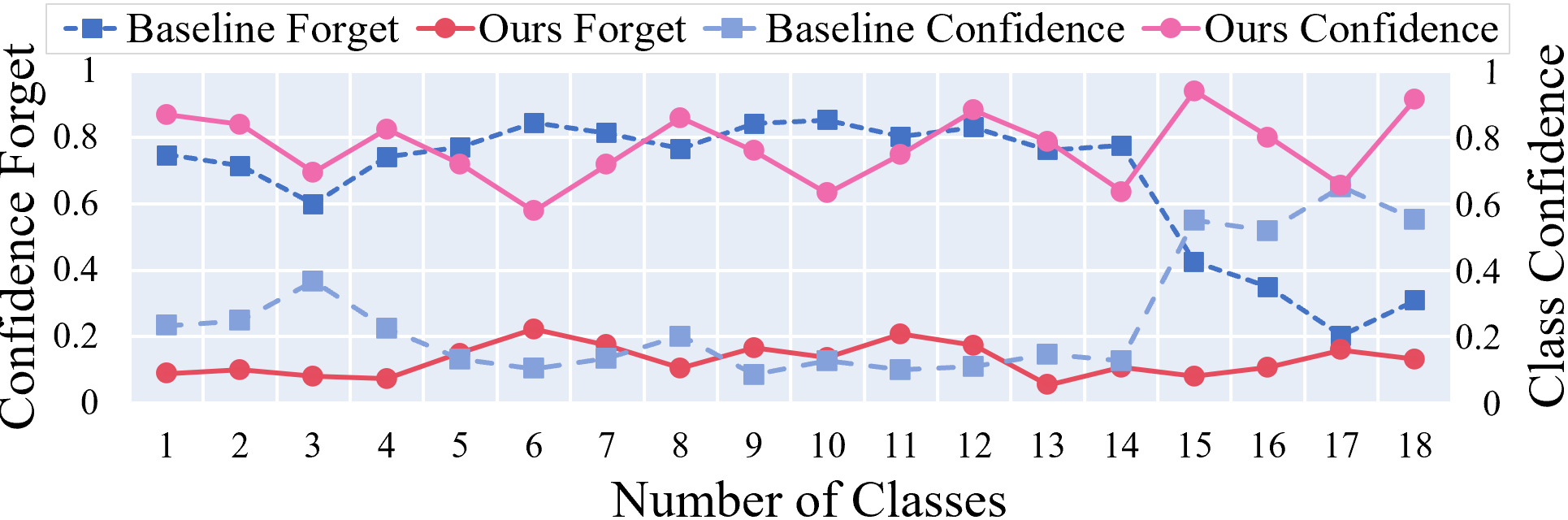}
  \caption{Confidence forgetting varies greatly among classes, making it difficult to effectively recall known knowledge by a unified and static pseudo-label threshold.}
  \label{fig:class-confidence}
\end{figure}

To illustrate this problem, we perform a quantitative analysis of the forgetting of class confidence which reflects how certain the model is about predictions. After performing incremental learning ($B10-C2$ protocol, explained in Experiments) on VOC datasets, we calculate the classification confidence distribution of each old class $p_{k}\sim \mathcal{N}(\mu_{k} ,\sigma^{2}_{k} )$, where $\mu_{k}$ is the mean confidence and $\sigma_{k}$ denotes variance:
\begin{align}
    \mu_k &= \frac{1}{|D_k|} \sum_{i=1}^{|D_{t-1}|} p_{ik} \cdot \mathds{1} (y_{ik} = 1),\\
    \sigma_k^2&=\frac{1}{|D_k|} \sum_{i=1}^{|D_{t-1}|} \left( p_{ik} \cdot \mathds{1} (y_{ik} = 1) - \mu_k \right)^2,
\end{align}

\noindent where $|D_{k}|$ is the number of samples where the true label for class $k$ is present. $p_{ik}$ is the predicted logit for class $k$ of the $i$-th sample. $\mathds{1}$ is the indicator function (1 when class $k$ is present, and 0 otherwise). Then, following the accuracy forgetting in SLCIL~\cite{chaudhry2018riemannian}, we define \textit{confidence forgetting} of each class as the difference between the maximum confidence gained throughout the past learning sessions and the confidence after finishing the current learning session: 
\begin{align}
    F_{k}=\max_{t\in \left \{ 1,\cdots ,T-1 \right \} } (\mu_{t,k} - \mu_{T,k} ),
\end{align}
where $\mu_{t,k}$ is the mean confidence of class $k$ on task $t$ when the model complete learning task $t$. As shown in Figure~\ref{fig:class-confidence}, the classification confidence of baseline suffers from severe forgetting up to 0.85, and the degree of forgetting varies greatly between classes, which makes it difficult to determine an optimal threshold for all classes. Here, we introduce a class-specific strategy to formulate different thresholds for each class based on the confidence distribution learned by the previous model.
According to the statistical principle, $3\sigma$ rule can guarantee the diversity of pseudo labels. For the representative examples of old classes, we define thresholds as mean confidence $\varepsilon_{k}=\mu_{k}$ and get the class-specific pseudo labels for training in session $t$:
\begin{equation}
    \hat{y}_{k}=\left\{\begin{matrix}
  1,& p_{k}\ge  \varepsilon_{k}, \\
  0,& p_{k}<  \varepsilon_{k}.
\end{matrix}\right.
\end{equation}
Since class distributions drift with sessions, we update distribution queues after completing each session to mitigate performance degradation caused by error accumulation.

\subsection{Probing Unknown Knowledge}
Previous works~\cite{song2023learning, zhou2022forward} have validated that learning extra classes at each session can reserve embedding space for future classes and enhance the model's forward compatibility. A straightforward way is to introduce real classes from other datasets as auxiliary, which is impractical due to access issues and potential domain discrepancies that may affect the current learning session. In multi-label scenarios, where images can encompass both known and latent unknown classes, we propose to mine knowledge in existing data to synthesize simulated features of future classes, thereby enriching the feature set. This prospective strategy pushes all features of the known classes presented in the current sample away from synthetic features, leading to a more compact and optimized representation of known classes, which better facilitates future learning. 

For an image in training data $D_{t}$, it has ground-truth label $Y=[y_{1},\cdots,y_{M}]\in\mathbb{R}^{M}$ where $M=|C_{1:t}|$, $y_{i}=1$ means $i$-th class exists in the image. Based on the dynamic feature purification, we obtain all class features $O_{S}\in \mathbb{R}^{|C_{1:t}|\times d}$, including features of known classes present and absent in current image, denoted as $O_{S1}=\left \{ o_{i}|y_{i}=1,o_{i}\in O_{S} \right \} \in \mathbb{R}^{M_{1}\times d}$ and $O_{S2}=\left \{ o_{i}|y_{i}=0,o_{i}\in O_{S} \right \} \in \mathbb{R}^{M_{2}\times d}$, respectively. The attention of present classes are localized to the fine-grained regions under label supervision, while attentions of other absent class features are freely distributed in other foreground regions. We leverage the implicit information in $O_{S2}$ to expand the model's feature set. Specifically, we randomly interpolate these absent class features to synthesize extra features $O_{V}\in \mathbb{R}^{1 \times d}$ as an unknown class:
\begin{align}
    O_{V}=\sum_{i=1}^{M_{2}} \bar{\lambda} _{i} \cdot o_{i},\bar{\lambda} _{i}=\lambda _{i}/{\textstyle \sum_{j=1}^{M_{2}}} \lambda _{j},
\end{align}
where $\lambda_{i}$ is randomly sampled from Beta distribution. Unknown features $O_{V}$ are then fed into the classifier jointly with real class features $O_{S}$ for classification results $P\in \mathbb{R}^{(M+1)}$. To this end, the original $M$-class problem is extended to a $(M+1)$-class. The label of unknown features is also a binary label to indicate whether it is real or synthetic. Combined with the stability and plasticity classifiers, we attach unknown classification to the plasticity classifier, which predicts $|C_{t}|+1$ classes. We must admit that the attention of absent classes may be paid to present classes due to inter-class similarity. However, the random synthetic features can simulate the distribution of future classes, and treating it as an additional class improves the model's ability to identify subtle differences, enhancing the compact representation of real classes and reserving space for future learning.    
\begin{table*}[t]
\small
\setlength{\tabcolsep}{1.5mm}
\begin{center}
\begin{tabular}{l|c|c|cccc|cccc}
\toprule
\multirow{3}{*}{\textbf{Method}} & \multirow{3}{*}{\begin{tabular}[c]{@{}c@{}}\textbf{Source}\\ \textbf{Task}\end{tabular}} & \multirow{3}{*}{\begin{tabular}[c]{@{}c@{}}\textbf{Buffer}\\ \textbf{Size}\end{tabular}} & \multicolumn{4}{c|}{\textbf{MS-COCO B0-C10}}                           & \multicolumn{4}{c}{\textbf{MS-COCO B40-C10}}                          \\ \cline{4-11} 
                        &                                                                        &                                                                        & \multicolumn{1}{c|}{Avg. Acc} & \multicolumn{3}{c|}{Last Acc} & \multicolumn{1}{c|}{Avg. Acc} & \multicolumn{3}{c}{Last Acc} \\ \cline{4-11} 
                        &                                                                        &                                                                        & \multicolumn{1}{c|}{mAP (\%)}      & CF1      & OF1      & mAP (\%)     & \multicolumn{1}{c|}{mAP (\%)}      & CF1      & OF1     & mAP (\%)     \\ \hline
Upper-bound             & Baseline                                                               & -                                                                      & \multicolumn{1}{c|}{-}        & 76.4     & 79.4     & 81.8    & \multicolumn{1}{c|}{-}        & 76.4     & 79.4    & 81.8    \\ \midrule \midrule
FT                      & Baseline                                                               & \multirow{7}{*}{0}                                                     & \multicolumn{1}{c|}{38.3}     & 6.1      & 13.4     & 16.9    & \multicolumn{1}{c|}{35.1}     & 6.0      & 13.6    & 17.0    \\
PODNet                  & SLCIL                                                                    &                                                                        & \multicolumn{1}{c|}{43.7}     & 7.2      & 14.1     & 25.6    & \multicolumn{1}{c|}{44.3}     & 6.8      & 13.9    & 24.7    \\
AGCN                     & MLCIL                                                                  &                                                                        & \multicolumn{1}{c|}{72.4}     & 53.9     & 56.6     & 61.4    & \multicolumn{1}{c|}{73.9}     & 58.7     & 59.9    & 69.1    \\
KRT                     & MLCIL                                                                  &                                                                        & \multicolumn{1}{c|}{74.6}     & 55.6     & 56.5     & 65.9    & \multicolumn{1}{c|}{77.8}     & {64.4}     & 63.4    & 74.0    \\
\textbf{HCP}                   & MLCIL                                                                  &                                                                        & \multicolumn{1}{c|}{\textbf{77.9}}         &  \textbf{60.4}        &  \textbf{65.3}        &  \textbf{71.2}& \multicolumn{1}{c|}{\textbf{78.9}}         &   \textbf{64.9}       & \textbf{68.6}        &  \textbf{75.3}\\ \midrule

TPCIL & SLCIL                                                                    & \multirow{6}{*}{5/class}                                               & \multicolumn{1}{c|}{63.8}     & 20.1     & 21.6     & 50.8    & \multicolumn{1}{c|}{63.1}     & 25.3     & 25.1    & 53.1    \\
PODNet& SLCIL                                                                    &                                                                        & \multicolumn{1}{c|}{65.7}     & 13.6     & 17.3     & 53.4    & \multicolumn{1}{c|}{65.4}     & 24.2     & 23.4    & 57.8    \\
DER++ & SLCIL                                                                    &                                                                        & \multicolumn{1}{c|}{68.1}     & 33.3     & 36.7     & 54.6    & \multicolumn{1}{c|}{69.6}     & 41.9     & 43.7    & 59.0    \\
AGCN & MLCIL                                                                  &                                                                        & \multicolumn{1}{c|}{72.9}     & 56.7     & 58.5     & 63.6    & \multicolumn{1}{c|}{74.5}     & 59.8     & 61.3    & 69.7    \\
KRT & MLCIL                                                                  &                                                                        & \multicolumn{1}{c|}{75.8}     & 60.0     & 61.0     & 68.3    & \multicolumn{1}{c|}{78.0}     & 66.0     & 65.9    & 74.3    \\
\textbf{HCP} & MLCIL                                                                  &                                                                        & \multicolumn{1}{c|}{\textbf{79.4}}         &   \textbf{70.3}       & \textbf{72.9}         &   \textbf{74.5}& \multicolumn{1}{c|}{\textbf{79.4}}         &   \textbf{71.5}       &  \textbf{74.1}       &  \textbf{76.7}\\ \midrule

iCaRL & SLCIL                                                                    & \multirow{9}{*}{20/class}                                              & \multicolumn{1}{c|}{59.7}     & 19.3     & 22.8     & 43.8    & \multicolumn{1}{c|}{65.6}     & 22.1     & 25.5    & 55.7    \\
BiC                     & SLCIL                                                                    &                                                                        & \multicolumn{1}{c|}{65.0}     & 31.0     & 38.1     & 51.1    & \multicolumn{1}{c|}{65.5}     & 38.1     & 40.7    & 55.9    \\
TPCIL                   & SLCIL                                                                    &                                                                        & \multicolumn{1}{c|}{69.4}     & 51.7     & 52.8     & 60.6    & \multicolumn{1}{c|}{72.4}     & 60.4     & 62.6    & 66.5    \\
PODNet                  & SLCIL                                                                    &                                                                        & \multicolumn{1}{c|}{70.0}     & 45.2     & 48.7     & 58.8    & \multicolumn{1}{c|}{71.0}     & 46.6     & 42.1    & 64.2    \\
DER++                   & SLCIL                                                                    &                                                                        & \multicolumn{1}{c|}{72.7}     & 45.2     & 48.7     & 63.1    & \multicolumn{1}{c|}{73.6}     & 51.5     & 53.5    & 66.3    \\
AGCN                  & MLCIL                                                                  &                                                                        & \multicolumn{1}{c|}{73.2}     & 59.5     & 60.3     & 66.0    & \multicolumn{1}{c|}{75.2}     & 64.1     & 65.2    & 71.7    \\
KRT                  & MLCIL                                                                  &                                                                        & \multicolumn{1}{c|}{76.5}     & 63.9     & 64.7     & 70.2    & \multicolumn{1}{c|}{78.3}     & 67.9     & 68.9    & 75.2    \\
\textbf{HCP}                     & MLCIL                                                                  &                                                                        & \multicolumn{1}{c|}{\textbf{79.6}}         &   \textbf{70.4}       &     \textbf{73.0}     &  \textbf{74.6}& \multicolumn{1}{c|}{\textbf{79.6}}         &          \textbf{71.9}&         \textbf{74.5}& \textbf{77.2}\\ \midrule \midrule

PRS                     & MLOIL                                                                  & \multirow{5}{*}{1000}                                                  & \multicolumn{1}{c|}{48.8}     & 8.5      & 14.7     & 27.9    & \multicolumn{1}{c|}{50.8}     & 9.3      & 15.1    & 33.2    \\
OCDM                    & MLOIL                                                                  &                                                                        & \multicolumn{1}{c|}{49.5}     & 8.6      & 14.9     & 28.5    & \multicolumn{1}{c|}{51.3}     & 9.5      & 15.5    & 34.0    \\
AGCN                   & MLCIL                                                                  &                                                                        & \multicolumn{1}{c|}{73.0}     & 59.4     & 65.9     & 59.0    & \multicolumn{1}{c|}{75.0}     & 63.1     & 64.8    & 71.1    \\
KRT                   & MLCIL                                                                  &                                                                        & \multicolumn{1}{c|}{75.7}     & 61.6     & 63.6     & 69.3    & \multicolumn{1}{c|}{78.3}     & 67.5     & 68.5    & 75.1    \\
\textbf{HCP}                    & MLCIL                                                                  &                                                                        & \multicolumn{1}{c|}{\textbf{79.5}}         &   \textbf{70.2}       &   \textbf{72.8}       &         \textbf{74.4}& \multicolumn{1}{c|}{\textbf{79.5}}         &   \textbf{71.8}       &  \textbf{74.4}       & \textbf{76.7}        \\ \bottomrule
\end{tabular}
\end{center}
\caption{Performance on MS-COCO, with comparison methods categorized by different source tasks. Buffer size 0 indicates no rehearsal is required, rendering many SOTA SLCIL approaches inapplicable. Best results among rows are highlighted in \textbf{bold}.}
\label{ms-coco-results}
\end{table*}

\subsection{Loss Function}
In long-term MLCIL tasks, pseudo-labels for previous classes rise with incremental sessions, leading to a disproportionate ratio of old to new class labels. Consequently, the model's pace of learning new classes tends to decelerate, which is particularly pronounced during the early training session. To sharpen the model's focus on new classes, we increase the weight of the classification loss for the new classes and adopt asymmetric loss (ASL). Given an image, we obtain its class probabilities $P=[p_{1},\cdots,p_{K}]\in\mathbb{R}^{K}$, where there are $K=|\mathcal{C} _{1:t}|+1$ classes including unknown class. The weighted ASL loss is as follows:
\begin{equation}
    L_{WASL} = \frac{1}{K} \sum_{k=1}^{K} w_{k}\cdot \left\{ \begin{array}{ll}
(1 - p_k)^{\gamma+} \log(p_k), & y_k = 1, \\
(p_k)^{\gamma-} \log(1 - p_k), & y_k = 0,
\end{array} \right.
\end{equation}
where $w_{k}$ is set to $\sqrt{\frac{|C_{1:t}|}{|C_{t}|} } $ for new classes and otherwise to 1. $y_{k}$ is the binary label to indicate whether label $k$ is present or not. $\gamma+$ and $\gamma-$ are used to manipulate the impact of positive and negative samples.

\section{Experiments}
\subsection{Experimental Setups}
\textbf{Datasets and Protocols.} HCP is evaluated on MS-COCO 2014~\cite{lin2014microsoft} and PASCAL VOC 2007~\cite{everingham2007pascal} datasets. MS-COCO contains 82,081 training images and 40,137 test images, which covers 80 common objects with an average of 2.9 labels per image. PASCAL VOC contains 5,011 images in the train-val set, and 4,952 images in the test set. It covers 20 common objects, with an average of 1.6 labels per image.
Similar to CIL works, we define different MLCIL protocols by a unified notation $Bi-Cj$, where $i$ denotes the number of classes learned in the base session and $j$ is the number of classes to be learned in each subsequent incremental session. We perform $B40-C10$ and $B0-C10$ protocols on MS-COCO dataset and $B10-C2$ and $B0-C4$ protocols on VOC 2007. We compare our method HCP with several baselines, representative SLCIL methods and state-of-the-art SLCIL methods with and without replay buffers.

\noindent \textbf{Evaluation Metrics.}
Similar to CIL, we adopt two widely used metrics for evaluation: average accuracy (Avg. Acc) and last accuracy (Last Acc). Following KRT, we use the mean average precision (mAP) to evaluate all the categories that have been learned in each session and report the average mAP (the average of the mAP of all sessions) and the last mAP (final session mAP). Two more metrics are all reported for a comprehensive multi-label performance evaluation, $i.e.$, the per-class F1 measure (CF1) and overall F1-measure (OF1) alongside the last accuracy.
\begin{table}[t]
\small
\setlength{\tabcolsep}{0.6mm}
\begin{center}
\begin{tabular}{l|c|cc|cc}
\toprule
\multirow{2}{*}{\textbf{Method}} & \multirow{2}{*}{\begin{tabular}[c]{@{}c@{}}\textbf{Buffer}\\ \textbf{Size}\end{tabular}} & \multicolumn{2}{c|}{\textbf{VOC B0-C4}} & \multicolumn{2}{c}{\textbf{VOC B10-C2}} \\ \cline{3-6} 
                        &                                                                        & Avg. Acc       & Last Acc      & Avg. Acc       & Last Acc      \\ \hline
Upper bound             & \multirow{2}{*}{-}                                                     & -              & 93.6          & -              & 93.6          \\
FT                      &                                                                        & 82.1           & 62.9          & 72.9           & 43.0          \\ \midrule
iCarL                   & \multirow{8}{*}{2/class}                                               & 87.2           & 72.4          & 79.0           & 66.7          \\
BiC                     &                                                                        & 86.8           & 72.2          & 81.7           & 69.7          \\
TPCIL                   &                                                                        & 87.6           & 77.3          & 80.7           & 70.8          \\
PODNet                  &                                                                        & 88.1           & 76.6          & 81.2           & 71.4          \\
DER++                   &                                                                        & 87.9           & 76.1          & 82.3           & 70.6          \\
KRT                   &                                                                        & 90.7           & 83.4          & 87.7           & 80.5          \\
\textbf{HCP}                      &                                                                        &  \textbf{93.5}          &    \textbf{89.2}           &       \textbf{92.1}         &  \textbf{86.3}             \\ \midrule
\textbf{HCP}  & 0    &   92.9          &    87.9 & 90.1    & 81.9  \\\bottomrule
\end{tabular}
\end{center}
\caption{Comparison results on PASCAL VOC dataset.}
\label{voc-results}
\end{table}

\noindent \textbf{Implementation Details.}
For fair comparison, we follow the experimental setting in KRT and use TResNetM~\cite{ridnik2021tresnet} pre-trained on ImageNet-21k~\cite{deng2009imagenet} as the backbone. We train the model with a batch size of 64 for 20 epochs, using Adam~\cite{kingma2014adam} optimizer and OneCycleLR scheduler with a weight decay of 1e-4. In the base session, we set the learning rate to 4e-5. In the following sessions, it adjusts to 1e-4 for MS-COCO and 4e-5 for VOC. In dynamic feature purification module, we set 3 attention blocks for VOC and 1 for MS-COCO. Our codes are available at https://github.com/InfLoop111/HCP.
\begin{table}[t]
\small
\setlength{\tabcolsep}{0.9mm}
\begin{center}
\begin{tabular}{ccc|cccccc|c}
\toprule
\multirow{2}{*}{FP} & \multirow{2}{*}{RE} & \multirow{2}{*}{PU} & \multicolumn{6}{c|}{Sessions}                 & \multirow{2}{*}{\begin{tabular}[c]{@{}c@{}}{Avg.}\\{Acc}\end{tabular}} \\ \cmidrule{4-9}
                    &                     &                     & 1     & 2     & 3     & 4     & 5     & 6     &                           \\ \midrule
                    &                     &                     & 97.58 & 89.10 & 86.30 & 61.31 & 56.16 & 46.76 & 72.87                     \\
                    \checkmark&                     &                     & 97.64 & 90.60 & 87.26 & 70.73 & 66.29 & 66.29 & 82.09                     \\
                    \checkmark&   \checkmark                  &                     &   97.84    &  94.17     &   90.72    & 83.44      &  76.39     & 73.45 & 86.00                     \\
                    \checkmark&                     &  \checkmark                   & 97.57 & 93.11&90.31&84.77&75.39&71.64 & 85.47                     \\
                    \checkmark&      \checkmark               &  \checkmark                   & 97.80 & 94.70  & 90.84 & 89.86  & 85.26  & 81.85 & \textbf{90.05 } \\ \bottomrule
\end{tabular}
\end{center}
\caption{Ablation study of each component. The module names are abbreviated as follows: FP-Feature Purification, RE-Recall Enhancement, PU-Probing Unknown.}
\label{ablation-component}
\end{table}
\begin{figure*}[t]
  \centering
  \includegraphics[width=0.92\linewidth]{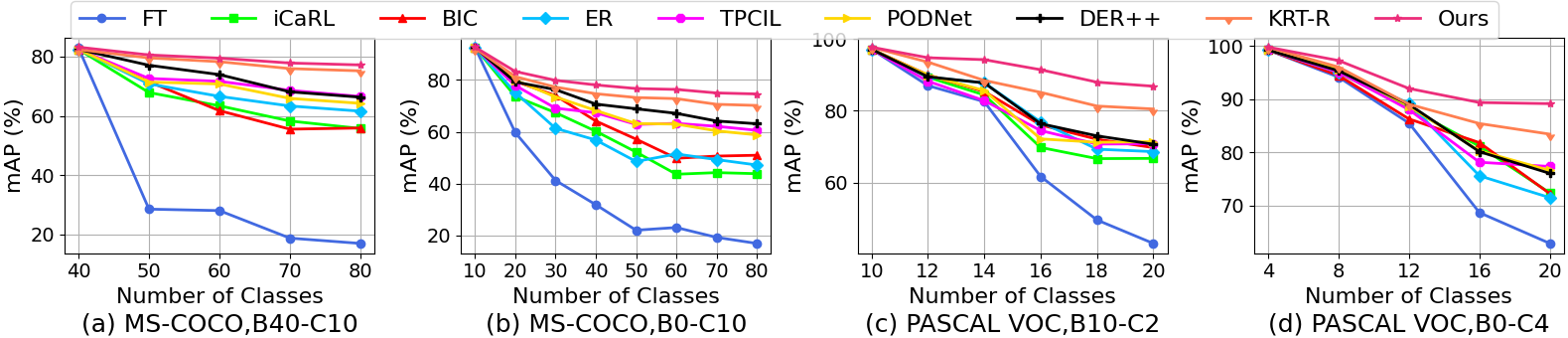}
  \caption{Performance curves (mAP\%) on MS-COCO and PASCAL VOC datasets under different protocols.}
  \label{curves_coco_voc}
\end{figure*}
\begin{table}[t]
\small
\setlength{\tabcolsep}{1mm}
\begin{center}
\begin{tabular}{c|cccccc|c}
\toprule
\multirow{2}{*}{\begin{tabular}[c]{@{}l@{}}{Recall}\\ {Strategy}\end{tabular}} & \multicolumn{6}{c|}{{Sessions}}                 & \multirow{2}{*}{\begin{tabular}[c]{@{}c@{}}{Avg.}\\ {Acc}\end{tabular}} \\ \cmidrule{2-7}
\multicolumn{1}{c|}{}                                                                                    & {1}     & {2}     & {3}     & {4}     & {5}     &{6}     &                                                                          \\ \midrule
$\varepsilon=0.8$                 & 97.85 & 92.79 & 90.40 & 78.22 & 70.01 & 68.94 & 83.08                                                                    \\
$\varepsilon=0.9$                 & 97.87 & 93.30 & 90.81 & 82.10 & 75.69 & 70.82 & 85.08                                                                    \\
Top-$2$               & 97.87 & 92.72 & 90.27 & 80.87 & 74.10 & 68.94 & 84.24                                                                    \\
\textbf{RE}                & 97.80 & 94.70  & 90.84 & 89.86  & 85.26  & 81.85 & \textbf{90.05}                                                                     \\ \bottomrule
\end{tabular}
\end{center}
\caption{Ablation of recall strategy, where the first two rows use a unified threshold, the third row represents top-K filtering, and RE is our Recall Enhancement.}
\label{tab: pseudo-ablate}
\end{table}

\subsection{Comparison Results}
\textbf{Results on MS-COCO.}
As shown in Table~\ref{ms-coco-results}, fine-tuning (FT) and SLCIL methods like PODNet suffer from severe forgetting in MLCIL tasks, with last accuracy 16.9\% and 25.6\% respectively, while our method achieves 71.2\% on B0-C10 when the buffer size is set to 0. Similarly, multi-label online incremental learning methods do not perform well, and our method outperforms PRS and OCDM by a large margin. Compared with SOTA MLCIL methods, our method still maintains a leading position, with improvements of up to 3.8\% in average accuracy over KRT (buffer size=1000) and a greater increase across all metrics compared to AGCN. Consistency improvements in B40-C10 setting underline the robustness of our HCP. It is notable that even without replay, our method exceeds all others with memory buffers in both average accuracy and last accuracy. Figure~\ref{curves_coco_voc}(a) and (b) show the performance curves as the number of classes increases. As incremental session progresses, our method exhibits stronger superiority, which demonstrates our effectiveness in long-term incremental scenarios.

\noindent \textbf{Results on PASCAL VOC.}
Table~\ref{voc-results} shows consistent improvements on VOC. For B0-C4 protocol, when buffer size is set to 0, our method outperforms KRT with 2/class replay in buffers by 2.2\% in Avg. Acc and 4.5\% in Last Acc respectively, which demonstrates the superiority of our method in scenarios with limited data access. HCP further achieves the best average mAP value of 93.5\% with extra buffers. For B10-C2 setting, HCP also exceeds other competitive methods by up to 4.4\% and 5.8\% and reaches the upper bound performance. Comparative curves in Figure~\ref{curves_coco_voc}(c) and (d) exhibit a widening gap between ours and other methods.
\subsection{Ablation Study}
\textbf{Effectiveness of Component.}
All ablation experiments are conducted on VOC B10-C2 setting. Tab.~\ref{ablation-component} reports results of Last Acc, Avg. Acc. Fine-tuning the old model with new data serves as baseline. After adding feature purification (FP), the model aggregates fine-grained features from multi-object images, and we get a 9.22\% Avg. gain. Recall Enhancement (RE) effectively recalls old known knowledge and alleviates forgetting difference among classes, which obtains a 13.13\% Avg. gain. For probing unknown knowledge, mining knowledge to generate features as unknown class enhances discriminability between real classes, boosting the performance of Avg. Acc by 12.60\%. Three modules jointly achieve the best, verifying the effectiveness of specifying what is known or not to accommodate historical, current and prospective knowledge.
\begin{figure}[t]
  \centering
  \includegraphics[width=0.98\linewidth]{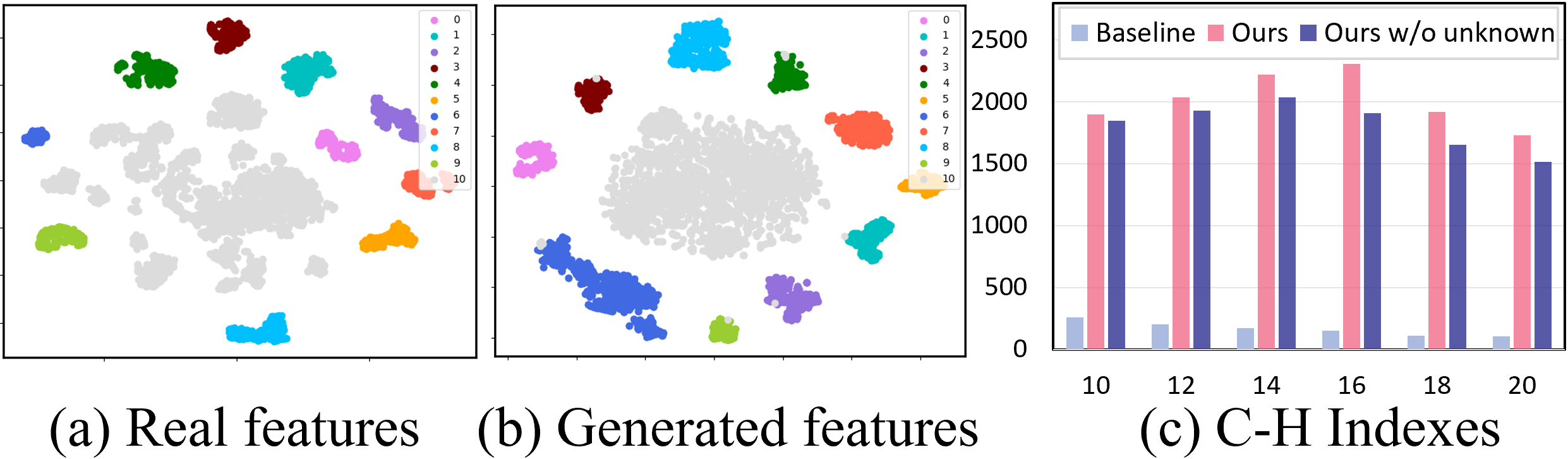}
  \caption{The distribution of (a) real future features can be simulated by (b) generated unknown features (gray in color). (c) Calinski-Harabasz (C-H) Indexes of class features.}
  \label{fig:virtual_index}
\end{figure}
\begin{figure}[t]
  \centering
  \includegraphics[width=0.97\linewidth]{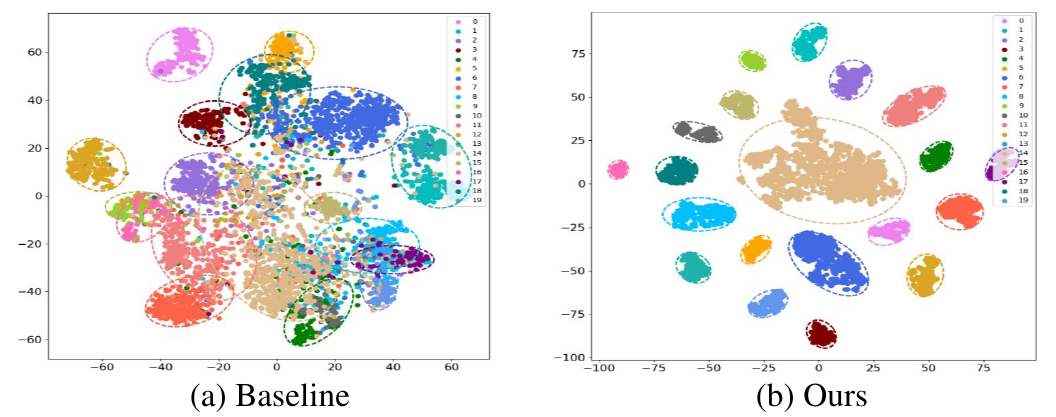}
  \caption{t-SNE visualization after incremental learning.}
  \label{fig:tsne_baseline_our}   
\end{figure}

\noindent\textbf{Ablation of Recall Strategy.} We compare different pseudo label selection strategies to recall past knowledge, including a unified threshold $\varepsilon$, top-$K$ filtering and enhancement with distribution prior. Table~\ref{tab: pseudo-ablate} shows that results are sensitive to the threshold. Our recall enhancement (RE), which considers the forgetting difference between classes and automatically determines the optimal threshold for each class, provides supervision with higher quality for old classes.

\noindent \textbf{Analysis of Feature Aliasing.} In Figure~\ref{fig:tsne_baseline_our}, we visualize the feature distributions of the baseline and our method after incremental learning on VOC dataset. It can be seen that the baseline method exhibits severe inter-class confusion and experiences catastrophic forgetting, while our method distinctly separates all classes without feature aliasing. 

\noindent \textbf{Analysis of Generated Unknown Features.} Figure~\ref{fig:virtual_index} (a) (b) compares the real future features and generated unknown features (gray in color), which both promote compact representation of current features. Besides, our interpolated features are very similar to the distribution of real future ones. This observation verifies the effectiveness of our probing unknown knowledge, which can provide valuable foresight for future learning. For quantitative illustration, we report the Calinski-Harabasz Index of feature representations in Figure~\ref{fig:virtual_index} (c), where a higher index indicates better separation between classes and compactness within classes. The index of ours at each session is significantly higher than the baseline. Moreover, without probing the unknown, the index drops a lot in the later sessions, which illustrates the advantage of virtual features for preparing the model for future learning.

\section{Conclusion}
In this paper, we present a novel method HCP for multi-label class-incremental learning, which specifies what is known or not at current learning session to accommodate historical, current and prospective knowledge. To clarify the known knowledge, feature purification is proposed to capture class-aware features from entangled global features, preventing feature aliasing within and between sessions. We analyze the confidence forgetting and further design recall enhancement to effectively retain historical known knowledge. To probe the unknown, we interpolate class features as prospective class to enhance the discriminative capacity and prepare for future learning. This provides a fresh insight into multi-label CIL problems. Comparisons with previous methods and our ablation study demonstrate the superiority of our overall design and the importance of each component in HCP.

\section{Acknowledgments}
\noindent This work is supported by the National Natural Science Foundation of China (Grant NO 62406318, 62376266,  62076195, 62376070),
 and by the Key Research Program of Frontier Sciences, CAS (Grant NO ZDBS-LY-7024).

\bibliography{main}

\end{document}